\documentclass[conference]{IEEEtran}
\IEEEoverridecommandlockouts{}

\def\arXiv{1}

\ifx\arXiv\undefined\else
\newcommand\copyrightnotice{%
\begin{tikzpicture}[remember picture,overlay]
\node[anchor=south,yshift=20pt] at (current page.south) {\fbox{\parbox{\dimexpr0.81\textwidth-\fboxsep-\fboxrule\relax}{\footnotesize \textcopyright \the\year{} IEEE. Personal use of this material is permitted. Permission from IEEE must be obtained for all other uses, in any current or future media, including reprinting/republishing this material for advertising or promotional purposes, creating new collective works, for resale or redistribution to servers or lists, or reuse of any copyrighted component of this work in other works.}}};
 \end{tikzpicture}%
}
\newcommand\acceptancenotice{%
\begin{tikzpicture}[remember picture,overlay]
\node[anchor=north,yshift=-15pt, text centered, text width=1.2\textwidth] at (current page.north) {\footnotesize This work has been accepted for publication at the 2025 IEEE International Automated Vehicle Validation Conference (IAVVC).};
\end{tikzpicture}%
}
\fi
\newcommand{\thetitle}{Improving AEBS Validation Through Objective Intervention Classification Leveraging the Prediction Divergence Principle}
\usepackage{amsmath,amsfonts}
\usepackage{algorithm}
\usepackage{amssymb}
\usepackage{array}
\usepackage{url}
\usepackage{verbatim}
\usepackage{graphicx}
\usepackage{csquotes}
\usepackage[%
  bibstyle=ieee,
  citestyle=numeric,
  isbn=true,
  doi=true,
  sorting=none,
  url=true,
  defernumbers=true,
  bibencoding=utf8,
  backend=biber
]{biblatex} 

\DeclareSourcemap{
  \maps[datatype=bibtex]{
     \map{
        \step[fieldsource=doi,final]
        \step[fieldset=isbn,null]
        \step[fieldset=issn,null]
        \step[fieldset=url,null]
        \step[fieldset=urldate,null]
        }
      }
}
\DeclareSourcemap{
  \maps[datatype=bibtex]{
     \map{
        \step[fieldsource=isbn,final]
        \step[fieldset=issn,null]
        \step[fieldset=url,null]
        \step[fieldset=urldate,null]
        }
      }
}

\usepackage{mathtools}
\usepackage[acronym]{glossaries}
\usepackage{tikz}
\usetikzlibrary{calc}
\usepackage{tikzscale}
\usepackage[]{xcolor}
\usepackage{tabularx} 
\usepackage{multirow}
\newcolumntype{C}{>{\centering\arraybackslash}X}
\usepackage{enumitem}
\usepackage[caption=false,farskip=0pt]{subfig}
\usepackage{pgfplotstable} 
\usepackage[hidelinks,
            bookmarks=false,
            pdfauthor={Daniel Betschinske; Steven Peters},
            pdftitle={\thetitle},
            pdfkeywords={Automatic Emergency Braking System (AEBS), False Positive (FP), Prediction Divergence Principle (PDP), Advanced Driver Assistance Systems (ADAS), Safety Validation, Classification, Pseudo Ground Truth, Field Operational Testing (FOT)}
            ]{hyperref}
\usepackage{orcidlink}
\makeatletter
\usepackage[top=0.75in,bottom=1.05in,left=0.625in,right=0.625in]{geometry}
\makeatother
\pgfplotsset{compat=1.16}

\newcommand{\figureref}[1]{Fig.~\ref{#1}}
\newcommand{\tableref}[1]{Table~\ref{#1}}
\newcommand{\sectionref}[1]{Section~\ref{#1}}

\tikzset{
    >=latex
}

\def\warningSign[#1,#2,#3]{%
  \pgfmathsetmacro{\scl}{#1}%
  \pgfmathsetmacro{\xpos}{#2}%
  \pgfmathsetmacro{\ypos}{#3}%
  \begin{scope}[shift={(\xpos,\ypos)}]

        \path[fill=red, rounded corners=0.3*0.1*\scl cm] (0,0) -- (0.3*1*\scl,0.3*1.732*\scl) -- (0.3*2*\scl,0) -- cycle;
        \pgfmathsetmacro{\inset}{0.78} 
        \begin{scope}[shift={(0.3*1*\scl,0.3*0.577*\scl)}, scale=\inset]
          \path[fill=white, rounded corners=0.3*0.07*\scl cm]
            ({0.3*-1*\scl},{0.3*-0.577*\scl}) -- (0.3*0,0.3*1.155*\scl) -- ({0.3*1*\scl},{0.3*-0.577*\scl}) -- cycle;
    
    \end{scope}
    \begin{scope}[shift={(0.3*1*\scl,0.3*0.577*\scl-0.06*\scl)}, scale=\inset*0.23]
            \filldraw[fill=black, draw=none, rounded corners=\scl*1pt]
            (-\scl*0.2,\scl*1.3) -- (\scl*0.2,\scl*1.3)  
            -- (\scl*0,\scl*0.25) 
            -- cycle;
        \fill (0,0) circle[radius=\scl*0.15];
    \end{scope}
  \end{scope}%
}

\newcolumntype{Y}{>{\centering\arraybackslash}X}
\newcolumntype{Z}{>{\raggedright\arraybackslash}X}
\newcolumntype{M}[1]{>{\raggedright\arraybackslash}m{#1}}

\pgfplotsset{every tick label/.append style={font=\footnotesize}}
\pgfplotsset{every legend/.append style={font=\footnotesize}}
\definecolor{bbox_edges}{RGB}{50,50,50}
\definecolor{egoObservedColor}{RGB}{255,165,0}

\definecolor{egoObservedColor}{RGB}{255,165,0}
\definecolor{objectObservedColor}{RGB}{0,0,255}

\definecolor{objectPredictedColor}{RGB}{0,255,255}
\definecolor{egoPredictedColor}{RGB}{255,255,0}
\definecolor{egoHypotheticalColor}{RGB}{0,255,0}
\definecolor{collisionColor}{RGB}{255,0,0}

\newacronym{FOT}{FOT}{field operational testing}
\newacronym{SOTIF}{SOTIF}{safety of the intended functionality}
\newacronym{SuT}{SuT}{system under test}
\newacronym{ODD}{ODD}{operational design domain}
\newacronym{GAMAB}{GAMAB}{globalement au moins aussi bon}
\newacronym{GAME}{GAME}{globalement au moins équivalent}
\newacronym{PRB}{PRB}{positive risk balance}
\newacronym{ALARP}{ALARP}{as low as reasonably possible}
\newacronym{MEM}{MEM}{minimal endogenous mortality}
\newacronym{MD}{MD}{minimum distance}
\newacronym{SBT}{SBT}{Scenario-Based Testing}
\newacronym{AD}{AD}{autonomous driving}
\newacronym{ADAS}{ADAS}{advanced driver assistance systems}
\newacronym{IS}{IS}{importance sampling}
\newacronym{AEBS}{AEBS}{automatic emergency braking system}
\newacronym[plural=$\mathrm{HBs}$, firstplural=hazardous behaviors]{HB}{$\mathrm{HB}$}{hazardous behavior}
\newacronym{EVT}{EVT}{extreme value theory}
\newacronym{PoT}{PoT}{peak over threshold}
\newacronym{DDT}{DDT}{dynamic driving task}
\newacronym{ASIL}{ASIL}{Automotive Safety Integrity Level}
\newacronym{E/E}{E/E}{electrical and/or electronic}
\newacronym{FuSa}{FuSa}{functional safety}
\newacronym{ADS}{ADS}{automated driving systems}
\newacronym{V&V}{V\&V}{verification and validation}
\newacronym{RAR}{RAR}{risk acceptance rationale}
\newacronym{EVA}{EVA}{extreme value analysis}
\newacronym{TPI}{TPI}{technical performance indicator}
\newacronym{IPM}{IPM}{intervention proximity metric}
\newacronym[plural=FPs, firstplural=false positives]{FP}{FP}{false positive}
\newacronym[plural=TPs, firstplural=true positives]{TP}{TP}{true positive}
\newacronym[plural=FNs, firstplural=false negatives]{FN}{FN}{false negative}
\newacronym[plural=TNs, firstplural=true negatives]{TN}{TN}{true negative}
\newacronym{FCPr}{FCPr}{false collision prediction}
\newacronym{TCPr}{TCPr}{true collision prediction}
\newacronym{CPr}{CPr}{collision prediction}
\newacronym{SuS}{SuS}{subset simulation}
\newacronym{XiL}{XiL}{x-in-the-loop}
\newacronym{PD}{PD}{Poisson distribution}
\newacronym{PDP}{PDP}{Prediction Divergence Principle}
\newacronym{CDF}{CDF}{cumulative distribution function}
\newacronym{PDF}{PDF}{probability density function}
\newacronym{MSE}{MSE}{mean squared error}
\newacronym{FoV}{FoV}{Field of View}
\newacronym{VRU}{VRU}{vulnerable road user}
\newacronym{TTC}{TTC}{Time to Collision}

\def\BibTeX{{\rm B\kern-.05em{\sc i\kern-.025em b}\kern-.08em
    T\kern-.1667em\lower.7ex\hbox{E}\kern-.125emX}}

\usetikzlibrary{external}
\begin{document}
\title{\thetitle}
\author{\IEEEauthorblockN{1\textsuperscript{st} Daniel Betschinske}
\IEEEauthorblockA{\textit{Institute of Automotive Engineering} \\
\textit{Technical University of Darmstadt}\\
Darmstadt, Germany \\
daniel.betschinske@tu-darmstadt.de\ifx\arXiv\undefined\else~\orcidlink{0000-0003-3203-2296}\fi}
\and
\IEEEauthorblockN{2\textsuperscript{nd} Steven Peters}
\IEEEauthorblockA{\textit{Institute of Automotive Engineering} \\
\textit{Technical University of Darmstadt}\\
Darmstadt, Germany \\
steven.peters@tu-darmstadt.de\ifx\arXiv\undefined\else~\orcidlink{0000-0003-3131-1664}\fi}
}
\maketitle
\ifx\arXiv\undefined\else
\copyrightnotice{}%
\acceptancenotice{}%
\fi
\begin{abstract}%
The safety validation of \gls{AEBS} requires accurately distinguishing between \gls{FP} and \gls{TP} system activations. %
While simulations allow straightforward differentiation by comparing scenarios with and without interventions, analyzing activations from open-loop resimulations --- such as those from \gls{FOT} --- is more complex. %
This complexity arises from scenario parameter uncertainty and the influence of driver interventions in the recorded data. %
Human labeling is frequently used to address these challenges, relying on subjective assessments of intervention necessity or situational criticality, potentially introducing biases and limitations. %
This work proposes a rule-based classification approach leveraging the Prediction Divergence Principle (PDP) to address those issues. %
Applied to a simplified \gls{AEBS}, the proposed method reveals key strengths, limitations, and system requirements for effective implementation. %
The findings suggest that combining this approach with human labeling may enhance the transparency and consistency of classification, thereby improving the overall validation process. %
While the rule set for classification derived in this work adopts a conservative approach, the paper outlines future directions for refinement and broader applicability. %
Finally, this work highlights the potential of such methods to complement existing practices, paving the way for more reliable and reproducible \gls{AEBS} validation frameworks.%
\end{abstract}

\glsresetall{}
\glsresetall{}
\section{Introduction}\label{sec::Introduction}
\Gls{ADAS} and \gls{ADS} aim to enhance road safety, comfort, and efficiency in road traffic~\cite{Tunnell.2018,Masello.2022, Neumann.2024} and are entrusted with increasingly complex driving functions in dynamic and partially unpredictable environments~\cite{Karunakaran.2022, Wang.2024}. %
A milestone in this evolution is the \gls{AEBS}, which is mandatory for newly registered cars in the European Union since July 2024~\cite{Regulation.2019_2144}. %
The intended functionality of \gls{AEBS} is to intervene in critical situations by applying brakes to prevent or mitigate collisions. %

One persistent challenge in \gls{AEBS} development is minimizing \gls{FP} brake events, which occur when the system intervenes unnecessarily~\cite{Lubbe.2014,Krishnan.2022}. %
\gls{FP} interventions are particularly problematic as they create risks in situations where no immediate hazard exists, potentially undermining safety benefits and eroding driver trust. %
Because simulations and proving-ground tests cannot fully reproduce real-world complexity~\cite{Tang.2023, Stocco.2023,Beringhoff.2022}, \gls{FOT} remains indispensable for uncovering unknown triggers for \gls{FP} and quantifying real-world performance~\cite{Krishnan.2022}. %
However, a fundamental challenge in evaluating real-world \gls{FP}-performance is the accurate classification of activations as \gls{TP} or \gls{FP}. %
Unlike controlled environments, where scenarios are designed to trigger specific system reactions, \gls{FOT} presents complex real-world conditions where sensor limitations and interventions by the \gls{SuT} or the driver make definitive classification challenging. %
Due to these challenges in classification, test engineers often rely on human judgment to categorize \gls{SuT} activations recorded during \gls{FOT} (e.g. \cite{Shaikh.2024}). %
This approach, however, introduces inherent subjectivity, as even with established criteria, human labelers may interpret the same traffic scenarios differently based on context~\cite{Kass.2018}, their background and experience~\cite{Crundall.2016}. %
Such subjective assessments can significantly undermine validation transparency, introduce systematic biases, and ultimately compromise the reliability of performance evaluations. %
Additionally, human labeling is resource-intensive and does not scale efficiently with increasing data volumes, limiting its applicability to more frequent events like \glspl{CPr} that do not trigger activations. %
To address these challenges, this paper proposes a novel rule-based classification approach leveraging the \gls{PDP}~\cite{Betschinske.2024}. %
Applied to a simplified \gls{AEBS}, this approach demonstrates its ability to complement existing validation practices by providing a conservative yet transparent framework for analyzing system activations.%

\section{Preliminaries}
As a foundation for the subsequent methodology, this section outlines the conventional classification schema of \gls{AEBS} responses and motivates the need for an alternative approach.%
\subsection{Classification of AEB System Activations}
In the context of \gls{AEBS} validation, system responses are typically classified into one of four main categories~\cite{Lubbe.2014, Krishnan.2022,Betschinske.2024, Helmer.2014, Knight.2022}: \acrfull{TP}, \gls{TN}, \acrfull{FP}, and \gls{FN}, as summarized in the confusion matrix in \tableref{tab:confusion_matrix}. %
\begin{table}
    \centering
    \caption{Confusion matrix for AEBS interventions adapted from~\cite{Betschinske.2024}}\label{tab:confusion_matrix}
    \begin{tabular}{|c|c|c|c|}
    \cline{3-4}
    \multicolumn{2}{c|}{} & \multicolumn{2}{c|}{\textbf{Triggered}} \\
    \cline{3-4} 
    \multicolumn{2}{c|}{} & \textbf{Positive} & \textbf{Negative} \\
    \hline
    \multirow{2}{*}{\textbf{Required}} 
      & \textbf{Positive} & True Positive (TP) & False Negative (FN) \\ \cline{2-4} 
      & \textbf{Negative} & False Positive (FP) & True Negative (TN) \\\cline{1-4} 
    \end{tabular}
\end{table}
\Gls{TP} and \gls{TN} classifications indicate correct system behavior, while \gls{FP} and \gls{FN} represent undesired responses. %
A \gls{TP} occurs when the system correctly initiates an intervention when required (use-case), thus mitigating risks and enhancing safety for vehicle occupants and other road users. %
Conversely, a \gls{TN} indicates that the system correctly abstains from intervention when it is not required, preserving normal vehicle operation. %
A \gls{FN} classification corresponds to situations in which the system fails to intervene despite an intervention being required. %
A \gls{FP} classification refers to unnecessary system interventions, where the \gls{AEBS} misperceives or misinterprets situations as requiring intervention, leading to unwarranted braking. %
Although \gls{FP} events do not necessarily cause accidents, they negatively affect the driving experience by causing discomfort, reducing user trust, and potentially increasing risks, such as rear-end collisions due to unexpected deceleration. %
Therefore, minimizing \gls{FP} interventions is crucial for both safety and user acceptance. %
\subsection{Interpretation of Intervention Requirement}
A key challenge in this process is defining when an intervention is \textit{required}. Three main interpretations exist in the literature:
The first interpretation is \textit{user-centered}, defining \textit{required} as interventions aligning with user expectations (e.g.,~\cite{Kallhammer.2011b}) treating deviations as failures. Due to the subjective nature of user expectations, this definition may yield varying interpretations of \gls{FP} and \gls{TP} depending on the user. %
The second interpretation is \textit{criticality-centered}, defining \textit{required} based on objective risk levels (e.g.,~\cite{Krishnan.2022,Helmer.2014, Lu.2021}). %
This approach assumes that criticality can be objectively quantified and evaluates the system's reaction against a criticality metric. %
However, \citeauthor{Helmer.2014} acknowledges that ``no generally accepted or universally applicable definition of 'dangerous' exists''~\cite{Helmer.2014}. %
The third interpretation is \textit{outcome-centered}, defining interventions as \textit{required} only when necessary to prevent collision. \citeauthor{Lubbe.2014}~\cite{Lubbe.2014} exemplifies this by classifying interventions as required only when collisions would otherwise be inevitable. %
\subsection{Problem Definition}
A fundamental issue is that none of the three interpretations provides a definitive criterion for evaluating intervention necessity, leaving assessment primarily dependent on the interpretation of human annotators, which is inherently subjective and context-dependent. %
Identical situations may receive different classifications based on context, outcomes, or individual perceptions, undermining consistent evaluation criteria. %
Further complications arise when labelers lack information about which object triggered the reaction of the \gls{SuT} or what prediction led to activation. %
For instance, human labelers might incorrectly classify an \gls{AEBS} reaction to a non-existent ``ghost object'' as appropriate if the overall situation appears critical for other reasons. %
Furthermore, human labeling is time-consuming and cost-intensive, limiting large-scale validation even for subcritical predictions. %
To address these challenges, this work implements and evaluates the \gls{PDP} approach proposed in~\cite{Betschinske.2024}, which offers an automated and more objective  classification method by evaluating the causal chain of system operations rather than relying solely on human interpretation. %

\section{Related Work}\label{sec::Related Work}
The evaluation of prediction accuracy through deviation analysis is well-established, typically involving comparisons between predicted states to a ground truth~\cite{Quehl.2018}. %
A notable example is presented in~\cite{Pink.29.03.2017}, where an ADAS function is monitored by continuously comparing predictions against updated predictions or sensor observations. %
If deviations exceed a predefined threshold, the system is deactivated. This approach exemplifies an operational realization of prediction divergence monitoring. %

The \gls{PDP} extends the concept of prediction divergence to the task of classification. %
It was originally defined in~\cite{Betschinske.2024} and is a fundamental concept for classifying decisions, interventions, or behavioral changes based on comparing the initial predicted outcome to a pseudo ground truth. %
This pseudo ground truth can be obtained using a more current prediction or the corresponding observation obtained at the time(s) to which the initial prediction referred. %

In this work, the concept of the \gls{PDP} is examined through a prototypical implementation aimed at classifying \gls{AEBS} interventions as either \gls{TP} or \gls{FP}. %
To the best of the author's knowledge, such an evaluation framework \textemdash{} particularly the construction of the pseudo ground truth as described in this paper \textemdash{} has not been previously presented in the literature. %

\section{Methodology}
\label{sec::Methodology}
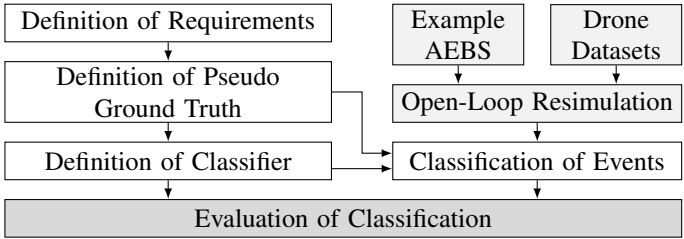
\begin{figure}
    \centering
    \begin{tikzpicture}[every node/.style={draw,rectangle,inner sep =2pt, minimum height =0.5 cm, text width = 4.2cm, text centered,anchor = north}, pdp/.style={fill=white!10},example/.style={fill=gray!10}]
        \def\verticalSpace{0.25cm}
        \node(Evaluation)[fill=gray!30, minimum width=\columnwidth ] {Evaluation of Classification};
        \node(Events)[anchor=south east,text width = 3.7cm] at ($(Evaluation.north east)+(0,\verticalSpace)$){Classification of Events};
        \node(Resim)[example, anchor=south east,text width = 3.7cm] at ($(Events.north east)+(0,\verticalSpace)$){Open-Loop Resimulation};
        \node(LevelXDatasets)[example, anchor=south east,text width = 1.6cm] at ($(Resim.north east)+(0,\verticalSpace)$){Drone Datasets};
        \node(AEBS)[example,anchor=south west,text width = 1.6cm] at ($(Resim.north west)+(0,\verticalSpace)$){Example AEBS};
        \node(Classifier)[pdp,anchor=south west]  at ($(Evaluation.north west)+(0,\verticalSpace)$){Definition of Classifier};
        \node(Pseudo)[pdp,anchor=south west]  at ($(Classifier.north west)+(0,\verticalSpace)$){Definition of Pseudo Ground Truth};
        \node(Requirements)[pdp,anchor=south west]  at ($(Pseudo.north west)+(0,\verticalSpace)$){Definition of Requirements};
        \draw[-latex] (Requirements) -- (Pseudo);
        \draw[-latex] (Pseudo) -- (Classifier);
        \draw[-latex] (Classifier.south) -- ++(0,-\verticalSpace);
        \draw[-latex] (AEBS.south) -- ++(0,-\verticalSpace);
        \draw[-latex] (LevelXDatasets.south) -- ++(0,-\verticalSpace); 
        \draw[-latex] (Resim.south) -- ++(0,-\verticalSpace);
        \draw[-latex] (Events.south) -- ++(0,-\verticalSpace);
        \draw[-latex] (Pseudo.east) -| ($(Pseudo.east)!0.5!(Events.west)$) |- ($(Events.north west)!0.3!(Events.south west)$);
        \draw[-latex] ($(Classifier.north east)!0.7!(Classifier.south east)$) -- ($(Events.north west)!0.7!(Events.south west)$);
    \end{tikzpicture}
    \caption{Methodology of this paper. The steps that implement the \gls{PDP} itself are colored white. The elements of the open-loop framework used for the subsequent evaluation of the classification are visualized in light gray.}
    \label{fig:Methodology}
\end{figure}

The methodology used for this paper is shown in \figureref{fig:Methodology}. %
First, the requirements for the classification of \gls{AEBS} activations are outlined. %
Then, a pseudo ground truth for \gls{CPr} is derived with respect to the requirements. %
Subsequently, a rule-based classification approach is presented, which allows for an automated classification of the \gls{AEBS} activations. %
To demonstrate the classification approach, a simplified \gls{AEBS} is implemented and open-loop simulated on several levelXData datasets \cite{highD.2018,exiD.2022,rounD.2020,inD.2020,uniD.2021}. %
Each time the \gls{AEBS} would have intervened, the current states, the predictions, the type of intervention, and a snippet containing the observations for the time scope of the predictions are extracted. %
Based on this data, the pseudo ground truth is derived, and the interventions are classified by applying a rule set.

The paper is structured as follows: \sectionref{sec::OpenLoopSimulation} details the simplified \gls{AEBS} implementation, which serves as the foundation for our evaluation methodology. %
Building upon this implementation, \sectionref{sec::PDP} elaborates on the core \gls{PDP} components highlighted in white in the methodology diagram. %
Together, these sections provide the foundation for the evaluation in \sectionref{sec::Evaluation}, where the \gls{PDP}-based classification approach is assessed against labels provided by human annotators. %

\section{Open-Loop Simulation Framework}\label{sec::OpenLoopSimulation}
This section presents the implementation of an open-loop simulation framework for generating \gls{AEBS} activations from real-world traffic recordings. These simulated activations serve as the basis for evaluating the classification approach proposed in \sectionref{sec::Evaluation}. %
\subsection{Simplified AEBS}\label{sec::ImplementationAEB}
\begin{figure}
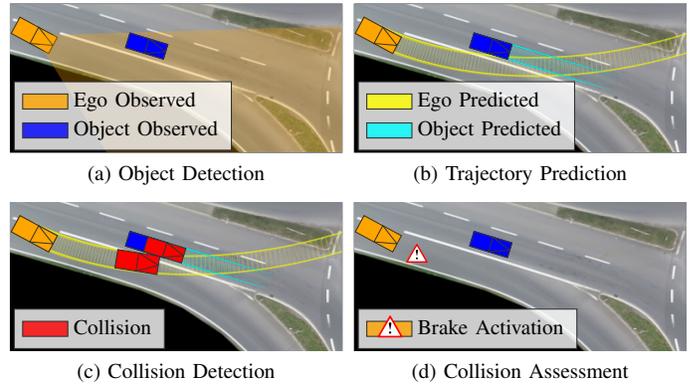

\subfloat[Object Detection]{%
\includegraphics[width=0.49\columnwidth]{figures/DroundR6F9601E234O222ObservedT0only.tikz}%
\label{fig:Pseudo:a}}%
\hfill
\subfloat[Trajectory Prediction]{%
\includegraphics[width=0.49\columnwidth]{figures/DroundR6F9601E234O222PredictedT0only.tikz}%
\label{fig:Pseudo:b}}%
\vspace{2mm}\\
\subfloat[Collision Detection]{%
\includegraphics[width=0.49\columnwidth]{figures/DroundR6F9601E234O222Collision.tikz}%
\label{fig:Pseudo:c}}%
\hfill
\subfloat[Collision Assessment]{%
\includegraphics[width=0.49\columnwidth]{figures/DroundR6F9601E234O222Activation.tikz}%
\label{fig:Pseudo:d}}%
\caption{Illustration of the output of the four subsequent modules of the simplified \gls{AEBS}.}%
\label{fig:AEBSModules}%
\end{figure}

The simplified \gls{AEBS} is designed to be lightweight, comprehensible, and efficient, rather than aiming for best performance. %
The system consists of four main modules visualized in \figureref{fig:AEBSModules}: the object detection, the trajectory prediction, the collision detection, and the collision assessment.
\subsubsection{Object Detection}
The \gls{AEBS} is designed to be active if the ego vehicle moves with a velocity greater than or equal to 1 m/s. Although only cars are equipped with the system in our simulation, the \gls{AEBS} can react to all types of traffic participants. 
To keep the implementation deterministic and efficient, the system omits concrete modeling of sensor setup or sensor fusion. %
Instead, the detection of entities is based on whether the center of the traffic participant is within the defined \gls{FoV} of the fictitious sensor setup of the \gls{AEBS}. %
The \gls{FoV} is parametrized with a $60^\circ$ view angle and a 200-meter range. %
As soon as a traffic participant is detected, the system receives access to all relevant information, such as position, velocity, and acceleration, available in the levelXData dataset. %
\subsubsection{Trajectory Prediction}
The trajectory prediction module employs a deterministic kinematic unicycle model with a 5~second horizon to predict the future states of the traffic participants based on their current position, orientation, translational and rotational velocities, and acceleration. %
The analytic model properly handles deceleration to standstill and uses temporal discretization matching the dataset sampling frequency of 25~Hz. %
For two-axle vehicles, the reference point is positioned at the estimated rear axle (85\% from the front bumper) to enhance cornering prediction accuracy. %
\subsubsection{Collision Detection}
The collision detection module identifies potential future collisions by systematically analyzing predicted trajectories. %
To reduce \glspl{FP} and improve efficiency, two types of objects are initially filtered out: %
those with an absolute sideslip angle above $12^\circ$, since such extreme values rarely occur in traffic, typically indicating errors in the dataset; %
and oncoming vehicles, which are recognized by comparing the heading and velocity vector of each object against those of the ego-vehicle. %
For the remaining entities, the module calculates the predicted \gls{MD} over the prediction horizon. %
When the \gls{MD} reaches zero within this horizon, a \gls{CPr} is generated. %
Each \gls{CPr} contains the information about the predicted ego and object trajectory, their predicted orientation, and dimensions. %
Additionally, the criticality is quantified by calculating the \gls{TTC}. %
If a \gls{CPr} is present in consecutive frames for more than 0.3 seconds, the \gls{CPr} is passed to the collision assessment module. %
\subsubsection{Collision Assessment}
The collision assessment module determines intervention timing and intensity using the \gls{TTC} as metric with two distinct intervention thresholds\footnote{These intervention thresholds are based on the PreSafe\textcopyright{} system~\cite{Bogenrieder.2009}, an early commercial \gls{AEBS} introduced in 2009 with full-braking capability.}: %
\begin{itemize}
    \item When the \gls{TTC} falls below 1.6 seconds, the system triggers a partial braking (moderate deceleration)
    \item When the \gls{TTC} falls below 0.6 seconds, an emergency braking is executed (maximum deceleration)
\end{itemize}
Both types of brake events are recorded during the simulation and serve as example events for the classification approach proposed in \sectionref{sec::PDP}. %
As no feedback is implemented, this procedure corresponds to an open-loop simulation of the \gls{AEBS}. %
\subsection{Datasets}\label{sec::Datasets}
Five datasets from levelXData were utilized for evaluation. %
These datasets are based on drone recordings of real-world traffic scenes, offering a top-down two-dimensional representation of trajectories, velocities, and accelerations of traffic participants, recorded at a frequency of 25 Hz. %
The datasets provide object metadata like the vehicle class, dimensions for non-\glspl{VRU}, and, except for highD, heading information. %
\tableref{tab:Datasets} provides an overview of the different domains and sizes of the datasets. %
\begin{table}[t]
\caption{Overview of levelXData datasets}\label{tab:Datasets}
\centering
\begin{tabularx}{\columnwidth}{|l|c|>{\raggedleft\arraybackslash}X|>{\centering\arraybackslash}m{0.8cm}|>{\centering\arraybackslash}m{0.8cm}|}
\hline
\multirow{2}{*}{\textbf{Dataset}} & \multirow{2}{*}{\textbf{Domain} }& \multirow{2}{1.4cm}{\centering\textbf{Distance traveled by cars}}  & \multicolumn{2}{c|}{\textbf{Evaluated Events}}\\\cline{4-5}
                                  &                                  &                                                                    &Partial Brake& Full Brake\\\hline
highD~\cite{highD.2018} & Highways           & 35961 km & 0   &  0  \\ \hline
exiD~\cite{exiD.2022}   & Highway Exits      & 20776 km & 0   &  14  \\ \hline
rounD~\cite{rounD.2020} & Roundabouts        & 1101 km  & 59  &  7 \\ \hline
inD~\cite{inD.2020}     & Intersections      & 602 km   & 150 &  19  \\ \hline
uniD~\cite{uniD.2021}   & University Campus  & 124 km   & 1   &  0  \\ \hline
\end{tabularx}
\end{table}
\subsubsection{Preprocessing}
To prepare the datasets for simulation, several preprocessing steps were implemented. %
The turn rate of the traffic participants was calculated using the gradient of the heading angle. %
Since highD lacks explicit heading data, the heading was estimated from velocity components under the assumption of negligible slip angles. %
To prevent calculation artifacts caused by the signal quantization and low-speed situations in highD, cubic spline smoothing was applied and headings interpolated for low-velocity cases ($\le 1$ m/s). %
Despite these preprocessing efforts, the highD dataset generated numerous \gls{CPr} but failed to trigger any AEBS activations. %
This can be attributed to the homogeneous traffic flow on highways, where situations with a \gls{TTC} below the intervention thresholds rarely occur which aligns with the conclusion in~\cite{Olleja.2022} that highD data is ``likely not suitable as an artificial source of critical events, neither with respect to timing (criticality) nor impact speed''. %
Finally, since the bounding box dimensions are not provided for \gls{VRU}, the minimum width of \gls{VRU} is set to 0.6 m, and the length is set to 1.5 m for motorcycles and bicycles and 0.6 m for pedestrians. %
\subsubsection{Simulation Results}\label{sec::SimResults}
A total of 358 brake events were generated by processing the datasets with the simplified \gls{AEBS} described above. %
In 108 events, the track of either the ego vehicle  or of the object traffic participant ended before the predicted time of collision, mostly due to the limited coverage area of the drones that recorded the dataset. %
Even though the classification criteria derived in \sectionref{sec::Classification} are robust with respect to disappearance --- since this would result in a conservative \gls{FP} classification --- these cases were excluded from the evaluation because there is no complete reference available for the human labelers. %

\section{Implementation of the PDP}\label{sec::PDP}
The following section presents the implementation of the \gls{PDP} approach, which enables the automated classification of \gls{AEBS} interventions. %
\subsection{Requirements for the Classification of AEBS Activations}\label{sec::Requirements} To support applicability in real-world \gls{FOT} and ensure the conservative evaluation required for safety validation, the approach must satisfy several key requirements, outlined below. %
\begin{enumerate}[
    leftmargin=*,
        label={R\arabic*},
        ref={R\arabic*}]
        \item The classification should be deterministic and free from subjective human judgment to ensure reproducible results.\label{R1}%
        \item The classification should be scenario-independent to ensure it works universally across all traffic situations, without requiring specific scenario detection algorithms that may introduce additional sources of error.\label{R2}%
        \item The classification method should be robust to variations in the sensor configuration of the \gls{SuT}, ensuring applicability to a wide range of \gls{AEBS} and enabling fair comparison between different systems.\label{R3}%
        \item The classification method should rely exclusively on data collected during the \gls{FOT}, accommodating the incompleteness of data.\label{R4} %
        \item The classification should compensate for influences from  driver actions or system interventions that alter the initial traffic situation.\label{R5}%
        \item The classification approach should be conservative, explicitly designed such that potential misclassifications only occur in the direction of labeling actual \gls{TP} activations as \gls{FP}, rather than the reverse to ensure safety-critical \gls{FP} are never mistakenly labeled as \gls{TP}.\label{R6} %
\end{enumerate}
\subsection{Pseudo Ground Truth}\label{sec::PseudoGroundTruth}
To retrospectively evaluate whether an intervention was  triggered correctly, the generated \gls{CPr} (see \sectionref{sec::ImplementationAEB}) is assessed against a defined pseudo ground truth. %
This pseudo ground truth builds upon the \textit{outcome-centered} definition by \citeauthor{Lubbe.2014}~\cite{Lubbe.2014}, classifying an intervention as a \gls{TP} if a collision would have occurred without it.%

We refine this definition in the context of \gls{AEBS} as follows: A \gls{CPr} is considered a \gls{TCPr} if, in the absence of longitudinal deceleration by the ego vehicle, a collision would have occurred; otherwise, it is considered a \gls{FCPr}.
Applying this definition directly requires assumptions since we cannot precisely determine what would have happened if the driver, or in closed-loop the system, had not intervened. %
Thus, we approximate the hypothetical trajectories of both the ego vehicle and surrounding traffic participants.%

For surrounding traffic participants (objects), we assume that their trajectories are unaffected by interactions with the ego vehicle during the evaluation time frame of the \gls{CPr}. %
This assumption is motivated by the  short prediction horizon of the \gls{AEBS} ($\leq 5 \mathrm{s}$) and its focus on frontal objects, making significant interactions within this time frame unlikely.%

For the ego vehicle, we assume that the driver continues along the initially intended path without significant evasive maneuvers. %
Under this assumption, the intended path is the observed one.
Although evasive actions occasionally occur, it is reasonable to assume drivers generally maintain their planned route within the short prediction horizon. 
However, in particular in scenarios where an imminent collision threat is present, the driver of the ego vehicle or the system are likely to intervene longitudinally. %
To mask these interventions, a hypothetical longitudinal driver behavior model is employed. %
This model maintains a constant acceleration from the instant of \gls{CPr} until the end of the evaluation interval to approximate a scenario in which the driver is inattentive to the potential collision object, which is the primary target situation for \gls{AEBS} interventions.%

\begin{figure}
    \centering
    \includegraphics[width=\columnwidth]{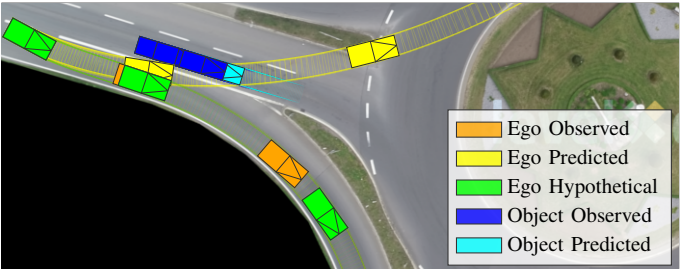}%
    \caption{Conceptual visualization of the hypothetical ego and the pseudo ground truth. The hypothetical ego is traveling with constant acceleration, as previously assumed in the prediction, along the observed path. The bounding boxes are shown at 0, 1.6 and 3.2 seconds. 
    The initial \gls{CPr} consists of the ego prediction and the object prediction. The pseudo ground truth consists of the hypothetical ego and the object observation.}\label{fig:Hyp}
\end{figure}

Thus, the hypothetical ego trajectory is derived by interpolating observed positional data with the hypothetical distance traveled under the assumption of constant acceleration. %
The concept of the hypothetical ego trajectory is illustrated in \figureref{fig:Hyp}. %
In conjunction with the observed trajectory of the object, this hypothetical trajectory serves as a pseudo ground truth for the \gls{CPr}, thereby enabling an objective assessment of the correctness and necessity of the \gls{AEBS} activation. %
\subsection{Rule-Based Classification Approach}\label{sec::Classification}

The \gls{PDP} states that the classification of prediction-based interventions can be determined based on the outcome of the prediction. %
Therefore, an intervention may be classified as correct given that the \gls{CPr} is correct and vice versa. %
Nonetheless, the classification of the \gls{CPr} requires the implementation of an appropriate evaluation metric and selection of a corresponding threshold that reflect the criterion that differentiates \glspl{TP} from~\glspl{FP}. %

While various prediction divergences, including absolute deviations between predicted and observed states, have the potential to serve as metrics for detecting system shortcomings, they do not directly reflect the necessity of interventions, thereby complicating the selection of appropriate thresholds. %
For instance, a small lateral deviation of a stationary object's predicted trajectory may significantly impact intervention necessity in scenarios involving close passes. %
In contrast, the same deviation in a car-following scenario has little influence, conflicting with the requirement for scenario independence~(\ref{R2}). %

In principle, the pseudo ground truth enables the evaluation of advanced criticality metrics, like those described in~\cite{Westhofen.2023}.\\*%
Still, defining generally acceptable deviation thresholds between the criticality of initial \gls{CPr} and the pseudo ground truth \gls{CPr} poses a significant challenge, since there are no established thresholds. %
While the \gls{MD} of the bounding boxes along the predicted and observed trajectories does not inherently quantify criticality, it is intrinsically linked to the necessity of an intervention. %
Specifically, the \gls{MD} in the pseudo ground truth \gls{CPr} reaching zero can be used as an indicator that a collision might have occurred without longitudinal intervention classifying the prediction as a \gls{TCPr}. %

It might seem intuitive to also classify \gls{CPr} as true if the \gls{MD} between the ego and the object is greater than zero but small --- for instance, closely passing a \gls{VRU}. However a relaxation of this criterion would lead to a non-conservative classification, and again, a potentially scenario-dependent threshold contradicting the requirements~\ref{R2} and~\ref{R6}. %

Thus, the adopted metric defines a \gls{CPr} as a \gls{TCPr} only if the \gls{MD} between the bounding boxes of traffic participants along the pseudo ground truth \gls{CPr} reaches zero. %
A notable disadvantage of this stringent criterion is its sensitivity to perceptual inaccuracies, as errors in observed bounding boxes at close proximity can falsely indicate collisions. %
Without a ground truth for those cases, the system would be incapable of determining whether the \gls{CPr} was correct or not. %
Consequently, it is imperative that collisions be meticulously documented within the \gls{FOT}. %
In the absence of such documentation, such predictions must be conservatively classified as \gls{FCPr} to meet~\ref{R6}. %

To summarize, this implementation of \gls{PDP}-based classification of \gls{CPr} is based on the following two conditions: %
\begin{equation}\label{eq::FCPr}
    \text{CPr} = \begin{cases}
        \mathrm{TCPr}, & \text{if }
        \left(\begin{aligned}
            \min_{t \in [0,T]} &\mathrm{MD}_{\mathrm{Pseudo}}(t) = 0 \\
            &\mathrm{AND}\\
            \min_{t \in [0,T]} &\mathrm{MD}_{\mathrm{Observed}}(t) > 0
        \end{aligned}\right) \\
        \mathrm{FCPr}, & \text{otherwise}
    \end{cases}
\end{equation}
where $\mathrm{MD}_{\mathrm{Pseudo}}$ and $\mathrm{MD}_{\mathrm{Observed}}$ denote the \gls{MD} between the ego vehicle and the object traffic participant in the pseudo ground truth and the observed data, respectively. %
Since the hypothetical ego and the object observation do not overlap in the pseudo ground truth in \figureref{fig:Hyp}, this \gls{CPr} is classified as a~\gls{FCPr}.

\section{Evaluation of the PDP-Classification}\label{sec::Evaluation}
This section evaluates the \gls{PDP}-based classification approach against human reference labels, analyzing annotation consistency and identifying potential directions for improvements.%
\subsection{Reference: Human Labeling}\label{sec::ReferenceClassification}
To assess the \gls{CPr} classifications, a reference labeling was created by three human labelers with a background in automotive engineering. %
Each labeler was instructed to review replays of the activations and evaluate them based on four selected criteria: %
\begin{enumerate}[
    leftmargin=*,
        label={Q\arabic*},
        ref={Q\arabic*}]
    \item From a drivers perspective: Would you wish for an intervention?\label{Q1}%
    \item Do you perceive the traffic situation as critical?\label{Q2}%
    \item Would you assume that, without a longitudinal intervention, a collision would have occurred?\label{Q3}%
    \item Would you label the event as a \gls{TP} (not FP)?\label{Q4}%
\end{enumerate}
Each question was rated on a five-point Likert-type scale~\cite{Likert.1932} ranging  from strong disagreement, somewhat disagreement, neutral/unsure, somewhat agreement, to strong agreement. %
The four questions were assessed by reviewing at least the top-view replay of the observation starting from the time of activation and lasting for a minimum of 5 seconds. %
However, labelers were allowed to watch a longer portion of the recording, including plots of the states of each traffic participant at their discretion if they felt it necessary for a more informed judgment. %
After completing those four questions, the predictions, the pseudo ground truth, and the \gls{CPr} classification were revealed. %
Subsequently, the labelers were instructed to answer a last question on the same scale, allowing the labelers to switch their decision compared to~\ref{Q4}:
\begin{enumerate}[
    leftmargin=*,
        label={Q\arabic*},
        ref={Q\arabic*}]
        \setcounter{enumi}{4}
    \item Do you agree with the \gls{CPr} classification of TCPr/FCPr?\label{Q5}%
\end{enumerate}
Finally, the labelers were instructed to annotate if they noticed a potential bug in the dataset. %

Note that the human annotators had access only to top-down replays without detailed contextual information, such as egocentric camera views or detailed driver behavior. %
Thus, their suitability as definitive ground truths is limited. %
Instead, the labels primarily serve as a qualitative reference for consistency checks and to identify potential for methodological improvement.
\subsection{Annotated Potential Bugs in Events}\label{sec::EvaluationOfBugs}
A total of 38 events were conservatively classified by the rule set as \gls{FCPr}, due to overlaps in the bounding boxes recorded during the event. %
Although the annotators were specifically instructed to flag such behaviors, the annotation coverage across these events varied between 50\% and 97\%.
Notably, one event with a bounding box overlap was missed by all three annotators. %
In three cases, at least one participant flagged an event as having a bounding box overlap, even though it was actually a close encounter without physical overlap. %
While this incorrect annotation does not impact the classification performance, these three events were conservatively excluded from the final evaluation to ensure consistency. %
Additionally, in five cases, at least one labeler annotated that the object exhibited implausible motion in the recording, such as unrealistically drifting trajectories or sudden high acceleration. %
These observations underscore the need to extend the approach with additional monitors to detect significant perception errors. %
If such perception errors are detected, the corresponding events should be conservatively flagged as \gls{FCPr} in accordance with requirement \hyperlink{R6}{R6}, unless a human expert explicitly overrules the classification. %
In return, the traffic participant perception and tracking of the \gls{SuT} have to be advanced enough that objects are tracked for a sufficient time and with adequate consistency to evaluate the \gls{CPr}, since prematurely vanishing objects and inconsistent tracking may increase the amount of \gls{FCPr} classifications. %

For the evaluation of the rule set, the 46 events annotated as containing potential bugs were excluded to prevent bias by data quality issues or annotation inconsistencies. %
This leaves a total of 204 events for subsequent evaluations. %
\subsection{Evaluation of the Reference}\label{sec::Reference}
To assess the reliability of the human reference ratings, the inter-rater agreement was quantified using Krippendorff's Alpha~\cite{Krippendorff.2019,Marzi.2024}. %
This metric accounts for agreement beyond chance and applies to ordinal scales and multiple raters. %
Krippendorff's Alpha ranges from -1 to 1, where 1 indicates perfect agreement, 0 indicates no agreement other than what would be expected by chance, and negative values indicate systematic disagreement. %
An Krippendorff's Alpha value above 0.80 is generally considered a strong agreement, while values between 0.67 and 0.80 indicate acceptable reliability. %

The Krippendorff's Alpha and the percentage of full agreement of the three labeling individuals per question are shown in \tableref{tab:Krippendorff}. %
\begin{table}[t]
    \caption{Krippendorff's Alpha and amount of full agreement for the answered questions}\label{tab:Krippendorff}
\begin{tabularx}{\columnwidth}{|>{\raggedright\arraybackslash}X|>{\centering\arraybackslash}m{0.2\columnwidth}|>
{\centering\arraybackslash}m{0.3\columnwidth}|}
\hline
 \textbf{Question} & \textbf{Krippendorff's Alpha} & \textbf{Same answer from all labelers} \\
\hline\ref{Q1} Situation Criticality & 0.711 & 69.1\% \\
\cline{1-3}\ref{Q2} Collision Likelihood & 0.655 & 72.1\% \\
\cline{1-3}\ref{Q3} Intervention & 0.642 & 70.6\% \\
\cline{1-3}\ref{Q4} Overall Label & 0.697 & 73.0\% \\
\cline{1-3}\ref{Q5} CPr Label Agreement & 0.120 & 76.0\% \\
\cline{1-3}\hline
\end{tabularx}
\end{table}
The observed differences across subjective evaluation criteria was moderate, with over two-thirds of activations receiving identical ratings from all three labelers. %

One likely explanation for the moderate inter-rater discrepancies, besides the subjectivity, is the limited visual and contextual information available during labeling. %
Human ratings were based solely on top-down replays of reconstructed bounding boxes, without access to egocentric camera views, traffic flow context, or behavioral responses of surrounding vehicles. %
This restriction complicates the judgment of criticality, collision likelihood, and the desirability of intervention, as well as the final overall label. %
Providing additional visual context (e.g., driver-view video recordings or scene annotations) could help improve inter-rater reliability in future evaluations. %
\subsection{Evaluation of the PDP-Classification}\label{sec::EvaluationClassification}
The analysis revealed a total of 36 \gls{TCPr} classifications and 168 \gls{FCPr} classifications, which were neither annotated by the rule set nor the labeler as a potential bug. %

\subsubsection{Overall Agreement}
To assess the overall agreement, deviations between the annotator ratings of~\ref{Q4} and the rule set classifications were analyzed. %
To facilitate a quantitative comparison, the qualitative Likert-type responses (from strong disagreement to strong agreement) were translated into numeric values (from 1 to 5). %
Since Likert-type scales are inherently ordinal and intervals between points are not necessarily equal, the numerical differences primarily indicate relative disagreement rather than exact measurements. %

Annotator deviations were calculated as absolute differences among their ratings, where neighboring answers result in a deviation of 1. %
To quantify the deviation compared to the \gls{CPr}-label two reference measures were used:
The first reference \textit{CPr-Q4} translated the binary rule set labels into extreme ratings, assigning ``strong agreement'' (5) to \gls{TCPr} and ``strong disagreement'' (1) to \gls{FCPr}. %
As the annotators were unaware of the \gls{CPr} labels  during the rating of~\ref{Q4}, deviations to this reference are unbiased. However, neutral answers are conservatively rated with a deviation of 2. %
The second reference \textit{CPr-Q5} directly measured annotators' subjective agreement with the rule set classification from~\ref{Q5}, ranging from 0 deviation (``strong agreement'') to 4 (``strong disagreement''). %
Both references indicate that annotators on average agree more with the rule set than with each other (see Table~\ref{tab:labelerdeviation}). %
This also holds for individual comparisons except for Annotator 3, who showed higher deviations to reference \textit{CPr-Q4} due to frequent use of neutral or ``somewhat'' responses (32 cases) compared to Annotators 1 (3 cases) and 2 (11 cases).
\begin{table}[tbp]
    \centering
    \caption{Average deviation between labelers and rule set}\label{tab:labelerdeviation}
    \begin{tabularx}{\columnwidth}{|>{\centering\arraybackslash}X|>{\centering\arraybackslash}X|>{\centering\arraybackslash}X|>{\centering\arraybackslash}X|>{\centering\arraybackslash}X|>{\centering\arraybackslash}X|}
    \cline{2-6}
     \multicolumn{1}{c|}{} & \textbf{Person\,1} & \textbf{Person\,2} &\textbf{Person\,3} & \textbf{CPr-Q4} & \textbf{CPr-Q5} \\
    \hline
    \textbf{Person\,1} & \textendash{} & 0.314 & 0.495 & 0.260 & 0.260 \\
    \hline
    \textbf{Person\,2} & 0.314 & \textendash{} & 0.377 & 0.152 & 0.103 \\
    \hline
    \textbf{Person\,3} & 0.495 & 0.377 & \textendash{} & 0.505 & 0.343 \\
    \hline
    \textbf{Average} & 0.404 & 0.346 & 0.436 & 0.306 & 0.235 \\
    \hline
    \end{tabularx}
\end{table}

\subsubsection{Analysis of Deviations}
The detailed results in \tableref{tab:Results} summarize the relative amounts of different agreements with the rule set obtained in~\ref{Q5}. %
The results are aggregated using the minimum, median, and maximum agreement ratings of the three labelers per event to reveal the spread of the agreement ratings. %
\begin{table}[b]
\caption{Aggregated agreement levels (in percent) for \gls{TCPr} and \gls{FCPr} events, based on the minimum, median, and maximum agreement ratings from the three annotators per event.}\label{tab:Results}
\begin{tabularx}{\columnwidth}{|>{\centering\arraybackslash}p{2.2 cm}|>{\centering\arraybackslash}p{2.3 cm}|>{\raggedleft\arraybackslash}X|>{\raggedleft\arraybackslash}X|}
\hline
 \multirow{2}{*}{\textbf{Label Aggregation}} & \multirow{2}{*}{\textbf{Agreement Level}}  &\multicolumn{2}{c|}{\textbf{PDP-Classification}} \\
\cline{3-4}
&&  TCPr & FCPr   \\
\hline
\multirow{5}{2.2 cm}{\centering \textbf{Minimum} agreement of the three labelers is selected} 
    & Strongly Disagree & 11.1\% & 6.0\% \\\cline{2-4}
    & Somewhat Disagree & 19.4\% & 1.8\% \\\cline{2-4}
    & Neutral           & 27.8\% & 0.6\% \\\cline{2-4}
    & Somewhat Agree    & 25.0\% & 3.0\% \\\cline{2-4}
    & Strongly Agree    & 16.7\% & 88.7\% \\
\hline
\multirow{5}{2.2 cm}{\centering \textbf{Median}\\ agreement of the three labelers is selected} 
    & Strongly Disagree & 2.8\%  & 0.0\% \\\cline{2-4}
    & Somewhat Disagree & 2.8\%  & 1.2\% \\\cline{2-4}
    & Neutral           & 0.0\%  & 1.2\% \\\cline{2-4}
    & Somewhat Agree    & 5.6\%  & 1.2\% \\\cline{2-4}
    & Strongly Agree    & 88.9\% & 96.4\% \\
\hline
\multirow{5}{2.2 cm}{\centering \textbf{Maximum} agreement of the three labelers is selected}
    & Strongly Disagree & 0.0\%  & 0.0\% \\\cline{2-4}
    & Somewhat Disagree & 0.0\%  & 0.0\% \\\cline{2-4}
    & Neutral           & 0.0\%  & 0.0\% \\\cline{2-4}
    & Somewhat Agree    & 0.0\%  & 0.6\% \\\cline{2-4}
    & Strongly Agree    & 100.0\%& 99.4\% \\
\hline
\end{tabularx}
\end{table}
This spread can partly be attributed to the subjectivity of the human labelers. This becomes especially apparent when comparing the agreement rates for \gls{TCPr}-classified events across different aggregation methods for the ratings of the three labelers: from 41.7\% (minimum) to 94.5\% (median) and 100\%~(maximum). %

An examination of the 21 \gls{TCPr}-classified events with a minimum label of neutral or lower agreement revealed two distinct patterns: %
First, in 18 cases, the ego vehicle was either still accelerating or had initiated light braking. %
In these scenarios, a small reduction in acceleration or a minimal increase in braking deceleration (significantly less than $1 \mathrm{m/s^2}$) would have prevented the collision. %
However, the \gls{TTC} evaluated on the pseudo ground truths showed only minor deviations from the~\gls{CPr}. %
Second, in 3 cases, the hypothetical ego vehicle would have intersected the observed object's trajectory significantly later ($> 1 \mathrm{s}$), indicating an overestimation of the situation's criticality. %
Both findings align with expectations, as the classification, as presented in this paper, is solely based on whether the pseudo ground truth collides within the prediction time frame, without considering the criticality of the pseudo ground truth itself. %
The quantification of the criticality in the pseudo ground truth by metrics such as the required acceleration or the \gls{TTC} and comparison with appropriate thresholds to evaluate correctness of the \gls{CPr} offers significant potential for addressing these cases, providing an even more conservative classification of \gls{CPr}. %

Even though the significantly higher amount of FCPr-classified events, only 15 FCPr-classified events received neutral or lower minimum agreement from at least one labeler. %
All these events involved scenarios where a \gls{VRU} or other traffic participant was passed with a small \gls{MD}. %
While such events can be identified automatically based on the pseudo ground truth, the threshold selection presents a more significant challenge. %
To maintain conservatism, a classification as \gls{FCPr} is recommended, until either well-trusted thresholds are established or until a human expert overrules the classification after thoroughly reviewing the event. %

\section{Conclusions}

This work introduces a prototypical, conservative, rule-based classification approach that leverages the \gls{PDP} to enhance the objectivity, reproducibility, and automation of \gls{AEBS} validation. %
By comparing the \gls{PDP}-based results with human annotations, we demonstrated that the method produces consistent classifications while reducing the impact of subjective variation in human labeling. %
These findings underline the potential of using an automated, objective rule set to enhance transparency, trust, and scalability in safety validation. %
However, until the identified improvements, such as the thresholds for overestimating criticality, are implemented, it is not recommended to replace human labeling with this approach entirely. %
Nevertheless, the rule set in its current form can already be integrated into the labeling process as pre-labeling support, an additional metric, or a consistency check to help human labelers and reduce subjectivity. %

\section{Outlook}
Future work should address the identified limitations by refining criticality metrics within the \gls{PDP}-based framework and investigating robust threshold selection to improve conservatism and accuracy further. Automatic methods for detecting and mitigating perception inaccuracies will also be essential to strengthen the reliability of the classification. In addition, combining \gls{PDP} with machine learning methods may help automate rule set selection and threshold tuning. %
In the long term, it may also be feasible to replace the rule set entirely with a machine learning-based approach that learns optimal classification boundaries directly from data. %
Given the limited dataset and the prototypical nature of the current implementation, it should be tested and refined with larger, more diverse datasets~\textemdash{}~ideally in collaboration with industry stakeholders~\textemdash{}~to ensure generalizability and prevent overfitting. Moreover, investigating the applicability of this approach to other \gls{ADAS} and automated driving systems could help establish broader use cases and promote standardization across driving functions. Finally, future research should explore initial steps toward standardizing \gls{PDP}-based validation approaches. Sharing best practices and results among research institutions and industry stakeholders might help improve the consistency and transparency of \gls{ADAS} evaluations.

\printbibliography{}

\end{document}